\documentclass[12pt]{article}

\usepackage[english]{babel}
\usepackage[T1]{fontenc}
\usepackage{carlito} 

\usepackage{newtxtext,newtxmath}

\usepackage{graphicx}
\usepackage{setspace}
\usepackage{cite}
\usepackage{amsmath,amsfonts}
\usepackage{algorithm}
\usepackage{algorithmic}
\usepackage{caption}
\usepackage{subcaption}
\usepackage{hyperref}
\usepackage{booktabs}
\usepackage{multirow}
\usepackage{enumitem}
\usepackage{adjustbox}
\usepackage{array}
\usepackage{titlesec}
\usepackage{geometry}
\usepackage{indentfirst}

\newcommand{\headingfont}{\rmfamily}

\titleformat{\section}
  {\headingfont\large\bfseries} 
  {\thesection}{1em}{}

\titleformat{\subsection}
  {\headingfont\normalsize\bfseries} 
  {\thesubsection}{1em}{}

\titlespacing*{\section}{0pt}{0.8em}{0.5em}
\titlespacing*{\subsection}{0pt}{0.6em}{0.4em}

\begin{document}

\begin{center}
{\headingfont\fontsize{18}{22}\selectfont\textbf{Graphing the Truth: Structured Visualizations for Automated\\
Hallucination Detection in LLMs}}
\end{center}

\begin{center}
\headingfont\textbf{Tanmay Agrawal}\\
Department of Computer Science, University of Arizona, Tucson, USA\\
tanmayagrawal21@arizona.edu
\end{center}

\begin{center}
\headingfont\textbf{\textit{Abstract}}
\end{center}

\doublespacing
\textit{Large Language Models have rapidly advanced in their ability to interpret and generate natural language. In enterprise settings, they are frequently augmented with closed-source domain knowledge to deliver more contextually informed responses. However, operational constraints such as limited context windows and inconsistencies between pre-training data and supplied knowledge often lead to hallucinations, some of which appear highly credible and escape routine human review. Current mitigation strategies either depend on costly, large-scale gold-standard Q\&A curation or rely on secondary model verification, neither of which offers deterministic assurance.
This paper introduces a framework that organizes proprietary knowledge and model-generated content into interactive visual knowledge graphs. The objective is to provide end users with a clear, intuitive view of potential hallucination zones by linking model assertions to underlying sources of truth and indicating confidence levels. Through this visual interface, users can diagnose inconsistencies, identify weak reasoning chains, and supply corrective feedback. The resulting human-in-the-loop workflow creates a structured feedback loop that can enhance model reliability and continuously improve response quality.}

\noindent
\textit{{\headingfont\textbf{Key Words:}} Hallucination Detection, Knowledge Graph Visualization, Retrieval-Augmented Generation, Human-in-the-Loop Auditing, Closed-Source Knowledge Integration}

\singlespacing

\section{Introduction}

Large Language Models (LLMs) have become central to modern natural language processing due to their strong performance in language understanding and generation \cite{Zubiaga2023-lg}. As organizations increasingly deploy these models in high-stakes, knowledge-centric environments, they often augment LLMs with closed-source datasets to improve factual grounding. While this improves contextual relevance, it also introduces new risks. Conflicts between pre-training distributions and domain-specific knowledge, combined with finite context windows, can lead LLMs to generate hallucinations that appear confident yet factually incorrect \cite{maynez-etal-2020-faithfulness,addlesee-2024-grounding,AnhHoang2025}.

Much of the current research targets hallucination prevention during generation through improved prompting, retrieval-augmented generation, or enhanced training pipelines. However, enterprise use cases frequently require post-hoc auditing of model outputs against authoritative sources, a capability that remains comparatively underdeveloped.

Recent work explores several post-hoc detection strategies, including evaluator-model frameworks, explainability-based inspection, and structured graph-based analysis \cite{Manakul2023-lw,Wiegreffe2019-tj,Sansford2024-if}. Systems such as GraphEval\cite{Sansford2024-if} demonstrate the promise of triplet-based comparisons for identifying factual inconsistencies, although limitations in scalability, transparency, and traceability persist. These challenges highlight the need for more interpretable and operationally efficient solutions.

Visualization techniques present an opportunity to address these gaps. Prior research shows that visual encodings of semantic relationships can support intuitive inspection, large-scale comparison, and human-in-the-loop validation of textual content \cite{wang-etal-2023-wizmap,Qiu2025-eh,hase-etal-2023-methods}. Motivated by these insights, this work proposes a visualization-driven framework that structures closed-source knowledge and LLM-generated statements as entity-relationship graphs. Through interactive visual exploration, the system enables users to identify potential hallucinations, trace their origins, and provide corrective feedback that supports more trustworthy and reliable deployment of LLMs in mission-critical applications.

\section{Literature Review}
Hallucination detection research follows three major directions. LLM-as-a-Judge approaches rely on a separate model to evaluate factual consistency, although they frequently inherit correlated biases and knowledge blind spots \cite{Manakul2023-lw}. Explainable AI techniques analyze model rationales, attention patterns, or generated explanations, but empirical evaluations indicate that many such explanations do not accurately capture the true generative process \cite{Wiegreffe2019-tj}. Graph-based techniques convert source material and model outputs into structured triplets or entity-relationship graphs, enabling deterministic comparisons and improved transparency. These methods have gained traction due to their reproducibility and interpretability \cite{Sansford2024-if}.

GraphEval represents a leading example of structured evaluation. It translates claims and source content into (subject, predicate, object) triplets and identifies factual mismatches through direct comparison \cite{Sansford2024-if}. Although effective, the method faces challenges involving computational overhead from Natural Language Inference pipelines, variability across NLI models, and limited traceability that forces users to manually inspect long prompt contexts. These limitations highlight the need for more interpretable and operationally efficient graph-based solutions.

Visualization-oriented research offers complementary insights into how complex textual relationships can be made more transparent. Systems such as WizMap position semantically related embeddings in close proximity and provide multi-resolution exploration of large knowledge spaces \cite{wang-etal-2023-wizmap}. VADIS uses a circular grid layout to simultaneously encode semantic similarity and query relevance, enabling intuitive navigation of document collections \cite{Qiu2025-eh}. Belief graph work by Hase et al. shows how node-edge representations can reveal relationships between model-stored factual assertions, supporting clearer inspection of inconsistencies \cite{hase-etal-2023-methods}.

Taken together, the literature points to two unresolved gaps: the need for structured and explainable mechanisms for hallucination detection, and the need for visualization frameworks that help users interpret relationships between generated claims and source knowledge. The present study addresses both gaps by integrating graph-based consistency checking with visualization-driven interpretability.

\section{Materials and Methods}{

\subsection{Dataset}
To evaluate the proposed approach, we used the SummEval dataset \cite{fabbri-etal-2021-summeval}, a widely adopted benchmark for assessing factual consistency in abstractive summarization. SummEval contains human evaluations of outputs from 16 summarization models applied to 100 CNN/DailyMail articles. Each generated summary is rated on a 1–5 Likert scale across four categories: consistency, coherence, fluency, and relevance. Building on prior work, including the TRUE benchmark \cite{Honovich2022-lo}, our study focuses specifically on the consistency dimension, where a score of 5 denotes complete factual alignment with the source document, and lower scores indicate varying degrees of inconsistency or hallucination.

The dataset comprises 1,600 annotated examples with a consistency-label ratio of 33.2\%. On average, model outputs contain 63 tokens, while source documents average 359 tokens, making SummEval an effective benchmark for evaluating hallucination detection methods under realistic summarization conditions.

\subsection{Structured Knowledge Representation}

We extract comprehensive entity-relationships from both closed-source
knowledge and LLM-generated outputs using triple extraction tech-
niques. Currently, we employ subject-verb-object triplets similar to
GraphEval \cite{Sansford2024-if}, but have extended this approach to apply extraction
methods to both source context and model outputs. While triple extrac-
tion offers a balance between simplicity and effectiveness, we continue
to explore richer representation formats including atomic sentences
with coreference resolution for capturing more complex relationships.

\subsubsection*{GraphEval+ Framework for Hallucination Detection}

We developed GraphEval+, an enhanced extension of the baseline GraphEval system, by integrating three primary improvements into the evaluation pipeline. First, the method performs bidirectional triple extraction, generating semantic triples from both the source material and the LLM output to ensure structurally comparable representations. Second, it introduces a semantic similarity matching stage that identifies the most relevant context triples for each generated triple, avoiding the inefficiency of comparing against the full prompt and mitigating known limitations of NLI models when handling long premises. Third, GraphEval+ employs targeted NLI evaluation, assessing each output triple only against its closest semantic matches from the source, resulting in a more focused and contextually grounded measure of factual consistency.

\subsubsection*{Alternative Approaches to Hallucination Detection}{
In addition to our primary GraphEval+ framework, we investigated alternative methodologies for detecting hallucinations in LLM-generated content through the Sentence Isolation plus CoReference Isolation (SICI) approach \cite{Xue2022-lo}. Instead of relying on LLM-driven triple extraction to convert text into subject-verb-object structures, SICI adopts a more direct representation by treating individual sentences as the fundamental units of analysis. The method resolves pronouns and entity references using Named Entity Recognition to ensure consistent entity tracking across sentences, and it incorporates variable contextual windows to support entity resolution where needed. Specifically, SICI-0 evaluates each sentence in isolation, while SICI-1 augments the evaluation by including one adjacent sentence before and after the target sentence to introduce minimal but meaningful contextual grounding. This streamlined design significantly reduces computational overhead, requiring roughly 30 minutes with T5 models compared to the 8-hour processing time for GraphEval+ executed with T5 and the OpenAI API, while still delivering competitive hallucination detection performance suitable for real-time or resource-constrained deployments.}

\subsection{ Visual Analytics Interface}

Our visualization approach builds on foundational principles demonstrated in systems such as WizMap \cite{wang-etal-2023-wizmap}, which shows how positioning semantically related embeddings in close proximity creates intuitive spatial representations of knowledge similarity, and VADIS \cite{Qiu2025-eh}, which uses a circular grid layout to communicate both semantic relatedness and query relevance. Extending these ideas, our interface positions output claim nodes according to their NLI consistency scores on the horizontal axis and their average semantic similarity to source claims on the vertical axis. Output claims are connected to their most relevant source claims through edges that encode semantic relationships, with shorter edges indicating stronger similarity. Color coding communicates reliability, with green representing highly reliable claims, orange indicating potentially suspicious content, and red flagging likely hallucinations. A force-directed layout prevents node overlap while preserving the semantic meaning of spatial arrangement. The visualization is organized into four interpretive quadrants: High Reliability, Suspicious Content, Plausible But Unsupported, and Potential Hallucination. Quantitative metrics, including average NLI score, semantic similarity score, and a combined confidence score weighted three-fourths NLI and one-fourth semantic similarity, offer rapid assessment of response quality. An interactive demonstration of this visualization is available online at \url{https://github.com/tanmayagrawal21/RAGChecker}.

\subsection{Evaluation}

To assess the effectiveness of our hallucination detection framework, we followed a structured evaluation protocol grounded in standard benchmarking practices. The evaluation was conducted using the ground-truth annotations provided in the SummEval dataset, which includes human-assigned consistency labels for model-generated summaries. These labels serve as the reference point for determining whether a given output is consistent with the source document or represents a hallucination.

For each example, the evaluation system first extracted the model-generated summary and its corresponding source article. The Hughes Hallucination Evaluation Model (HHEM)\cite{Sansford2024-if} was applied to compute a factual consistency score based on semantic and inferential alignment between the summary and the source. We treated HHEM as a baseline and then incorporated our proposed methods, which augment or replace portions of the evaluation pipeline with structured graph representations and SICI-based contextual inference techniques.

All approaches were assessed using the Balanced Accuracy Score, defined as the average of recall across both the consistent and hallucinated classes. This metric was selected because hallucination detection often exhibits class imbalance, where purely accuracy-based reporting may mask poor performance on minority classes. By using balanced accuracy, we obtain a more equitable assessment of model performance across both correct and incorrect factual claims.

Each evaluation method was executed on the same dataset split to ensure comparability. Processing time was also recorded to quantify the computational efficiency of each approach. Together, these steps provide a standardized and repeatable framework for measuring hallucination detection performance and for isolating the incremental contribution of each methodological component.}

\section{ Results and Discussions }
Our visualization framework provides an interpretable, spatial representation of hallucination behavior in Large Language Model outputs. As shown in Figure 1, each generated claim is positioned within a two-dimensional space defined by its NLI consistency score and semantic similarity to the source material. 

\begin{figure}[t]
    \centering
    \includegraphics[width=0.9\linewidth]{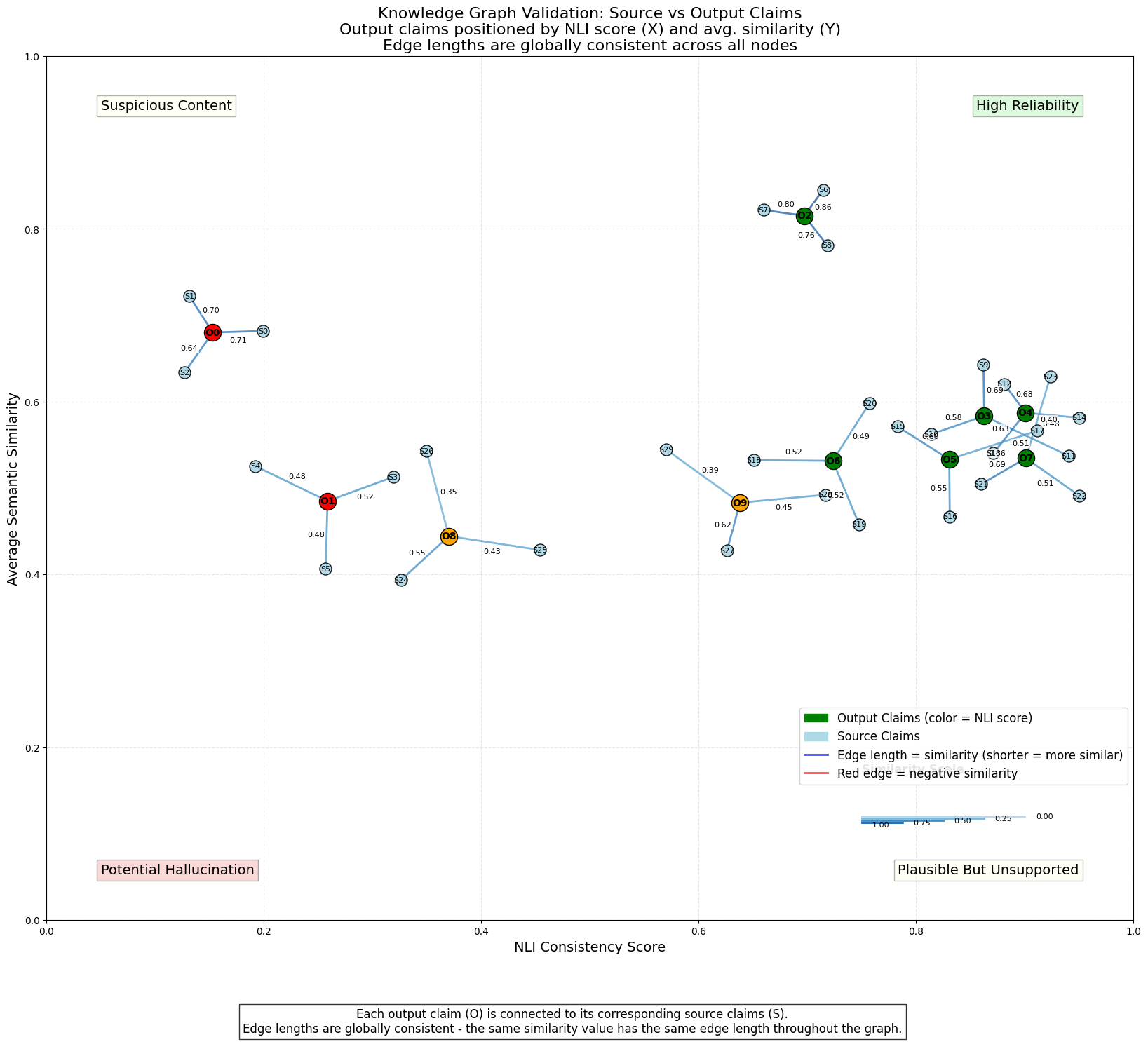}
    \caption{Visualization approach showing claim reliability through spatial positioning and color coding.}
    \label{fig:fig1}
\end{figure}

Figure 2 illustrates how the quadrant-based layout and color encoding support rapid identification of reliable content, suspicious statements, unsupported claims, and likely hallucinations. By adapting visual grammars from prior work in graph-based belief mapping, such as Hase et al. \cite{hase-etal-2023-methods}, the system enables users to trace generated statements back to their most relevant source relationships and understand how factual inconsistencies emerge.

\begin{figure}[t]
    \centering
    \includegraphics[width=0.9\linewidth]{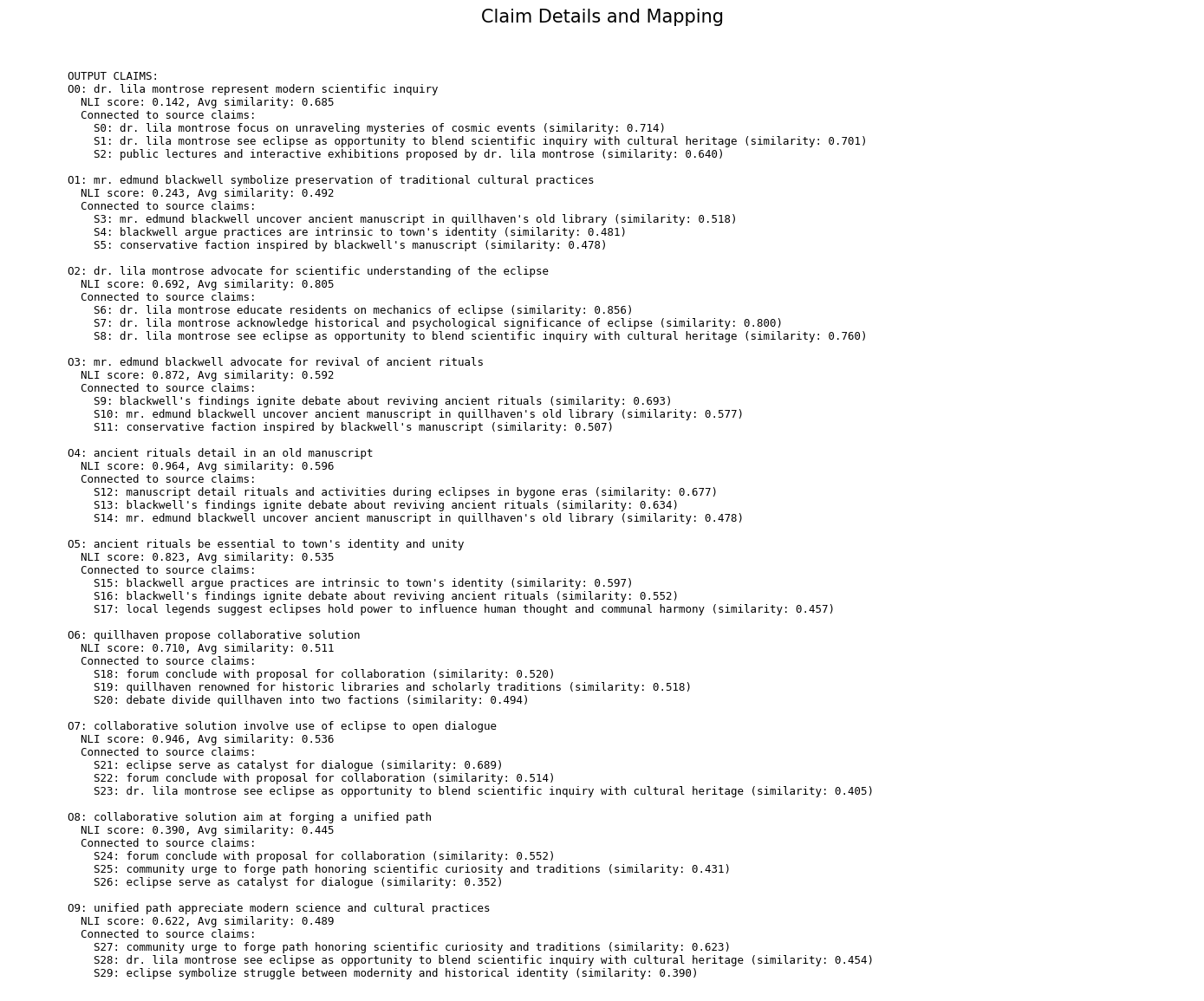}
    \caption{Visualization metadata showing the interpretation of different quadrants and color codes.}
    \label{fig:fig2}
\end{figure}

To quantitatively assess hallucination detection performance, we evaluated several methodology variants using Balanced Accuracy as the primary metric. Table 1 summarizes results on the SummEval dataset. The baseline HHEM model reached 66\%, while the original GraphEval approach improved performance to 71.5\%. 

\begin{table}[t]
\centering
\caption{Hallucination detection performance on SummEval}
\begin{tabular}{l c l}
\toprule
\textbf{Method} & \textbf{Accuracy} & \textbf{Notes} \\
\midrule
HHEM \cite{Sansford2024-if} & 66.0 & Baseline from GraphEval paper \\
HHEM+GraphEval \cite{Sansford2024-if} & 71.5 & From GraphEval paper \\
HHEM+GraphEval+ & 53.0 & $\sim$8h (T5 \& OpenAI API) \\
HHEM+SICI-0 & 67.0 & $\sim$30min (T5) \\
HHEM+SICI-1 & 70.0 & Using T5 model \\
\bottomrule
\end{tabular}
\label{tab:summeval}
\end{table}

In contrast, our enhanced GraphEval+ method achieved 53\%, indicating that additional structural complexity did not translate into improved detection accuracy. A likely contributing factor is that GraphEval+ categorizes any LLM failure in triple extraction as a hallucination, which can inflate false-negative rates and reduce balanced accuracy. Notably, the SICI approaches offered competitive performance with far lower computational costs. SICI-0 achieved 67\% accuracy, while SICI-1 reached 70\%, nearly matching GraphEval performance at a fraction of the runtime.

These findings highlight important trade-offs across hallucination detection methodologies. Graph-based systems remain highly effective for structured comparison and visual analysis, yet may incur increased computational overhead and sensitivity to extraction errors. In contrast, the SICI approaches provide strong accuracy-efficiency balance, with SICI-1 demonstrating the value of even minimal contextual windows for entity-level reasoning. From a deployment standpoint, the considerable runtime difference between GraphEval+ (approximately eight hours) and SICI-based methods (approximately thirty minutes) underscores the relevance of computational constraints in practical workflows.

The visualization component further complements these quantitative results by supporting human-centered verification workflows. Users can rapidly identify problematic claims through the quadrant organization, examine source-claim linkages via semantic edges, and iteratively refine outputs through the interactive interface. This approach reduces cognitive load by preserving accurate content while isolating unreliable elements, avoiding the need to discard entire responses. Moreover, users gain actionable insight into knowledge gaps when clusters of outputs appear in the "Plausible but Unsupported" region, prompting updates to source documentation. Early feedback from domain reviewers emphasizes that the system's confidence-spectrum view aligns more closely with real human decision-making than binary verification tools, suggesting value in enhancing transparency and trust in LLM-assisted environments.

Overall, the results demonstrate that structured visual analytics offers a complementary dimension to automated hallucination detection. While not always maximizing classification accuracy, the visual framework provides interpretability, transparency, and human-in-the-loop capabilities that are essential for evaluating and improving factual reliability in real applications.

\section{ Conclusion and Future Works}

This work demonstrates that structured visual analytics can substantially enhance the detection and interpretation of hallucinations in large language model outputs. By mapping claims into a spatial framework grounded in NLI consistency and semantic similarity, the system shifts verification from a binary correctness check to a graded assessment of reliability. This approach reduces cognitive load for reviewers, distinguishes between different categories of problematic content, and supports selective retention of accurate statements rather than discarding entire responses.

Empirical results highlight key tradeoffs across evaluation methods. While GraphEval+ introduces methodological sophistication through bidirectional extraction and targeted consistency checks, it achieved lower balanced accuracy compared to the original GraphEval framework. This outcome underscores a critical insight that increasing methodological complexity does not always yield performance gains, especially when extraction failures are interpreted as hallucinations. Conversely, the simpler SICI-1 method achieved competitive accuracy with far greater computational efficiency, reinforcing the value of lightweight techniques for scalable, automated hallucination detection. Together, these findings suggest that diverse methodological pathways remain necessary, depending on whether the goal is accuracy, efficiency, or interpretability.

Future work will focus on increasing interactivity within the visualization, enabling users to manipulate claim positions, explore node-level details on demand, and engage in an iterative refinement workflow where updated responses are automatically re-evaluated. As LLMs continue to influence high-stakes decision environments, transparent verification mechanisms will become increasingly critical. The proposed visualization-driven framework contributes toward this goal by making the relationship between generated claims and source knowledge comprehensible, navigable, and conducive to human oversight in contexts demanding factual dependability.

\bibliographystyle{IEEEtran}
\bibliography{bibtex/bib/IEEEexample}

@ARTICLE{Zubiaga2023-lg,
  title    = "Natural language processing in the era of large language models",
  author   = "Zubiaga, Arkaitz",
  journal  = "Front. Artif. Intell.",
  volume   =  6,
  pages    = "1350306",
  year     =  2023,
  language = "en"
}

@inproceedings{addlesee-2024-grounding,
    title = "Grounding {LLM}s to In-prompt Instructions: Reducing Hallucinations Caused by Static Pre-training Knowledge",
    author = "Addlesee, Angus",
    editor = "Dinkar, Tanvi  and
      Attanasio, Giuseppe  and
      Cercas Curry, Amanda  and
      Konstas, Ioannis  and
      Hovy, Dirk  and
      Rieser, Verena",
    booktitle = "Proceedings of Safety4ConvAI: The Third Workshop on Safety for Conversational AI @ LREC-COLING 2024",
    month = may,
    year = "2024",
    address = "Torino, Italia",
    publisher = "ELRA and ICCL",
    url = "https://aclanthology.org/2024.safety4convai-1.1/",
    pages = "1--7"
}

@article{AnhHoang2025,
  title = {Survey and analysis of hallucinations in large language models: attribution to prompting strategies or model behavior},
  volume = {8},
  ISSN = {2624-8212},
  url = {http://dx.doi.org/10.3389/frai.2025.1622292},
  DOI = {10.3389/frai.2025.1622292},
  journal = {Frontiers in Artificial Intelligence},
  publisher = {Frontiers Media SA},
  author = {Anh-Hoang,  Dang and Tran,  Vu and Nguyen,  Le-Minh},
  year = {2025},
  month = sep 
}

@inproceedings{maynez-etal-2020-faithfulness,
    title = "On Faithfulness and Factuality in Abstractive Summarization",
    author = "Maynez, Joshua  and
      Narayan, Shashi  and
      Bohnet, Bernd  and
      McDonald, Ryan",
    editor = "Jurafsky, Dan  and
      Chai, Joyce  and
      Schluter, Natalie  and
      Tetreault, Joel",
    booktitle = "Proceedings of the 58th Annual Meeting of the Association for Computational Linguistics",
    month = jul,
    year = "2020",
    address = "Online",
    publisher = "Association for Computational Linguistics",
    url = "https://aclanthology.org/2020.acl-main.173/",
    doi = "10.18653/v1/2020.acl-main.173",
    pages = "1906--1919"
}

@ARTICLE{Sansford2024-if,
  title        = "{GraphEval}: A {Knowledge-Graph} based {LLM} hallucination
                  evaluation framework",
  author       = "Sansford, Hannah and Richardson, Nicholas and Maretic,
                  Hermina Petric and Saada, Juba Nait",
  year         =  2024,
  primaryClass = "cs.CL",
  eprint       = "2407.10793"
}

@article{fabbri-etal-2021-summeval,
    title = "{S}umm{E}val: Re-evaluating Summarization Evaluation",
    author = "Fabbri, Alexander R.  and
      Kry{\'s}ci{\'n}ski, Wojciech  and
      McCann, Bryan  and
      Xiong, Caiming  and
      Socher, Richard  and
      Radev, Dragomir",
    editor = "Roark, Brian  and
      Nenkova, Ani",
    journal = "Transactions of the Association for Computational Linguistics",
    volume = "9",
    year = "2021",
    address = "Cambridge, MA",
    publisher = "MIT Press",
    url = "https://aclanthology.org/2021.tacl-1.24/",
    doi = "10.1162/tacl_a_00373",
    pages = "391--409"
}

@ARTICLE{Honovich2022-lo,
  title        = "{TRUE}: Re-evaluating factual consistency evaluation",
  author       = "Honovich, Or and Aharoni, Roee and Herzig, Jonathan and
                  Taitelbaum, Hagai and Kukliansy, Doron and Cohen, Vered and
                  Scialom, Thomas and Szpektor, Idan and Hassidim, Avinatan and
                  Matias, Yossi",
  year         =  2022,
  primaryClass = "cs.CL",
  eprint       = "2204.04991"
}

@ARTICLE{Qiu2025-eh,
  title        = "{VADIS}: A visual analytics pipeline for dynamic document
                  representation and information-seeking",
  author       = "Qiu, Rui and Tu, Yamei and Yen, Po-Yin and Shen, Han-Wei",
  year         =  2025,
  primaryClass = "cs.HC",
  eprint       = "2504.05697"
}

@inproceedings{wang-etal-2023-wizmap,
    title = "{W}iz{M}ap: Scalable Interactive Visualization for Exploring Large Machine Learning Embeddings",
    author = "Wang, Zijie J.  and
      Hohman, Fred  and
      Chau, Duen Horng",
    editor = "Bollegala, Danushka  and
      Huang, Ruihong  and
      Ritter, Alan",
    booktitle = "Proceedings of the 61st Annual Meeting of the Association for Computational Linguistics (Volume 3: System Demonstrations)",
    month = jul,
    year = "2023",
    address = "Toronto, Canada",
    publisher = "Association for Computational Linguistics",
    url = "https://aclanthology.org/2023.acl-demo.50/",
    doi = "10.18653/v1/2023.acl-demo.50",
    pages = "516--523"
}

@inproceedings{hase-etal-2023-methods,
    title = "Methods for Measuring, Updating, and Visualizing Factual Beliefs in Language Models",
    author = "Hase, Peter  and
      Diab, Mona  and
      Celikyilmaz, Asli  and
      Li, Xian  and
      Kozareva, Zornitsa  and
      Stoyanov, Veselin  and
      Bansal, Mohit  and
      Iyer, Srinivasan",
    editor = "Vlachos, Andreas  and
      Augenstein, Isabelle",
    booktitle = "Proceedings of the 17th Conference of the European Chapter of the Association for Computational Linguistics",
    month = may,
    year = "2023",
    address = "Dubrovnik, Croatia",
    publisher = "Association for Computational Linguistics",
    url = "https://aclanthology.org/2023.eacl-main.199/",
    doi = "10.18653/v1/2023.eacl-main.199",
    pages = "2714--2731"
}

@ARTICLE{Xue2022-lo,
  title        = "{CorefDRE}: Document-level Relation Extraction with
                  coreference resolution",
  author       = "Xue, Zhongxuan and Li, Rongzhen and Dai, Qizhu and Jiang,
                  Zhong",
  year         =  2022,
  primaryClass = "cs.CL",
  eprint       = "2202.10744"
}

@ARTICLE{Manakul2023-lw,
  title        = "{SelfCheckGPT}: Zero-resource black-box hallucination
                  detection for generative Large Language Models",
  author       = "Manakul, Potsawee and Liusie, Adian and Gales, Mark J F",
  year         =  2023,
  primaryClass = "cs.CL",
  eprint       = "2303.08896"
}

@ARTICLE{Wiegreffe2019-tj,
  title        = "Attention is not not Explanation",
  author       = "Wiegreffe, Sarah and Pinter, Yuval",
  year         =  2019,
  primaryClass = "cs.CL",
  eprint       = "1908.04626"
}

\end{document}